\documentclass[letterpaper]{article}
\usepackage{amsmath}
\usepackage{amsmath}
\usepackage{amsfonts}
\usepackage{amssymb}
\usepackage{graphicx} 
\usepackage{cite}
\usepackage{multirow}
\usepackage{url}
\urldef{\email1}\path|{a.ter-sarkisov,s.r.marsland}@massey.ac.nz|
\title{Convergence of a Recombination-Based Elitist Evolutionary Algorithm on the Royal Roads Test Function}
\begin{document}
\author{Aram Ter-Sarkisov \ Stephen Marsland\\
 Department of Computer Science\\
 Massey University, New Zealand\\
 \email1\\
\url{http://www.massey.ac.nz/seat}}
\maketitle
\begin{abstract}
We present an analysis of the performance of an elitist Evolutionary algorithm using a recombination operator known as 1-Bit-Swap on the Royal Roads test function based on a population. We derive complete, approximate and asymptotic convergence rates for the algorithm. The complete model shows the benefit of the size of the population and recombination pool.   
\end{abstract}
\section{Introduction}
Evolutionary Algorithms (EA) are a set of heuristic optimization tools, that are well-suited to problems with poorly-understood landscapes (sometimes known as black-box optimization). Despite a rich history in application, theoretical analysis has been lagging behind; while there have been some advances in recent years, the analysis has mainly been restricted to single-parent algorithms, which is an extremely limiting assumption. 

The algorithm that we analyse is described as a $(\mu + \lambda)$ evolutionary algorithm, where $\mu$ and $\lambda$ are the size of the population from which solutions are picked, and the size of the mating pool that is created from them, respectively. The $+$ signifies that the algorithm is elitist, i.e., that the best solution at each iteration is always reproduced in the next, so that the best fitness in the population can never decrease. In this paper we consider the effect that population has on the expected time when the algorithm finds the optimal solution to the problem. We model the distribution of elite species in the population using a static model of a uniform distribution. Most work to date has considered only $(1+1)$ EAs. The operator that we use to recombine solutions is the 1-bit-swap (1BS) operator that was described in~\cite{tersarkisov2010}.  

We derive an exact expression for the expected runtime of $(\mu+\lambda)$EA$_{1BS}$. Since this expression does not appear to have a closed form, we then develop an approximation to it and compute the asymptotic limit. Rather surprisingly, in the asymptotic expression the size of the population and recombination pool cancel out, and so they do not appear.

\subsection{Royal Roads Function}

The Royal Roads (RR) is a test function introduced in \cite{mitchell92} and analyzed in \cite{mitchell96}, where a population-based Evolutionary Algorithm was found to have underperformed a simpler heuristic Randomized Local Search (RLS), which contradicted the theoretical findings in the same article. The Royal Road was initially developed to demonstrate the schemata theory of EAs. A string of length $n$ is split into $K$ consecutive bins or blocks (which we index as $\kappa_1, \kappa_2,\ldots, \kappa_K$). All bins have the same size, $M$, so that $n=KM$. The fitness of the string is the sum of the fitness for each bin, and the fitness of each bin is $M$ if all the bits in the bin have value 1, and 0 otherwise. Thus, the possible fitness of strings with the Royal Road function are $0, M , 2M, \ldots, KM$, so the function has plateaus of fitness, where a large number of strings have the same fitness value. This means that single bit modifications to the solution string will generally not improve the fitness function. This means that within the plateau, the solutions created by the EA are expected to show a random walk behaviour. 

\subsection{Past Work}
RR has not received much attention in recent EA literature, where the focus has been on the rather simpler OneMax (also known as Counting Ones) fitness function. However, in \cite{storchwegener03} an upper bound on the expected running time of $O(n^6)$ was found for a version of RR. In \cite{mitchell96} the bounds on convergence for RR were found to be $O(2^K \log n)$, where $n$ is the length of the string and $K$ the length of the bin, up to a linear term tighter than the bound for the RLS ($O(2^K N \log N)$), although numerically RLS outperformed EA. This result does not involve the size of the population or recombination pool in any way.

For the OneMax test function there has been rather more research on EAs, although $(1+1), (\mu+1) \textnormal{ and } (1+\lambda)$ set-ups are still more widespread. In \cite{heyao02} it was shown that the effect of population is problem-specific, i.e., increase in population size may not improve performance at all. Very recently, in \cite{chen11}, it was shown that populations of size $O(\log n)$ boost performance, while those of size $\Omega(\frac{n}{\log n})$ impair the progress of the algorithm (with the analysis based on the TrapZeros multimodal function) and reduce the probability of global convergence.

\section{Analyzed Algorithm:$(\mu+\lambda)$EA$_{1BS}$}
\label{sec:aa}
The genetic operator that we consider in this paper is not the usual mutation operator. The k-Bit-Swap genetic operator (KBS) was introduced in \cite{tersarkisov2010}. It contains some features of both mutation and uniform crossover and recombines information between two parents in a random manner. In this article we use 1-Bit-Swap (1BS), which picks exactly 1 bit from each parent uniformly at random.\\       
\linebreak
See Table \ref{tab:tab1} for the pseudocode of the algorithm.\\
\linebreak
Tournament selection consists of picking two species at random and putting the fitter of the pair into the mating pool.\\
\linebreak
\begin{table}
\begin{tabular}{|r|l|}
\hline
1&Initialize population size $\mu$\\
& $\quad$ repeat for t generations:\\
2& $\quad \quad$ select $\frac{\lambda}{2}$ pairs of parents from the population using Tournament selection\\
& $\quad \quad $ repeat $\frac{\lambda}{2}$ times:\\
3a& $\quad \quad \quad $ select a bit at random in Parent 1\\
3b& $\quad \quad \quad $ select a bit at random in Parent 2\\
3c& $\quad \quad \quad $ swap values in the selected bits\\
4& $\quad \quad $ after the recombination, keep $\alpha$ best species in the population, \\
& $\quad \quad $ replace the rest with the best species from the pool\\
\hline
\end{tabular}
\caption{$(\mu+\lambda)$EA$_{1BS}$}
\label{tab:tab1}
\end{table}

\section{Model Setup and Assumptions}
\label{sec:model}
The main quantity we analyze in this article is the first hitting time of the global solution of the test problem:
\begin{equation*}
\mathbf{\tau}^{RR}_A=\min\{t \geq 0:f(\alpha)=n\}
\end{equation*}  
where $A$ is the set of all possible populations that include a global solution. We want to find $\mathbf{E}\tau^{RR}$, the expectation of this time parameter, for the $(\mu + \lambda)$ EA with 1BS as the only genetic operator.
 
\subsection{Improvement process}
We start with the pessimistic assumption that each bin $\kappa$ starts with an equal number of 0s and 1s, which implies that the starting fitness of all elements of the population is 0. As the Royal Road fitness function makes incremental improvements impossible to see, in order to measure the progress of the algorithm we introduce, in addition to the fitness function, an auxiliary function, in this case OneMax (where the fitness of a string is simply a count of the number of 1s in it; for further reference see e.g. \cite{chenhe09}). We denote the value of the auxiliary function for a bin $\kappa$ as $V_{\kappa}$, which can theoretically have values between 0 and $M$, although in practice they all start at $M/2$ because of our assumption above. Since only 1 bit is changed in a string at each iteration, only one bin can evolve at a time. In a slight abuse of notation, we refer to that bin as the `active bin' and index it as $\kappa$. Within a bin, the number of improvements that have already been made (i.e., the number of bits that were 0 and have already been changed into 1s) is denoted by the variable $l$, which starts at $0$ and increases to $M/2$.

We restrict our attention to elite pairs in the recombination pool, i.e. pairs in which both parents are currently elite species. This is a limitation of our analysis that means that we underestimate the chance of success. The probability of selecting an elite pair in the recombination pool is:
\begin{equation*}
P_{\mathrm{sel}}(\alpha)=\frac{\alpha^2(\alpha+2(\mu-\alpha))^2}{\mu^4}=\frac{(\alpha(2 \mu- \alpha))^2}{\mu^4}   
\end{equation*}
\noindent where $\alpha$ is the number of elite species in the population. 

Having selected the pair, the probability that as a results of swapping bits between them, a better species evolves is: 
\begin{equation*}
P_{\mathrm{swap}}=\frac{2(\frac{M}{2}-l)(\frac{n}{2}+\frac{\kappa M}{2}+l)}{n^2}=\frac{(M-2l)(n+\kappa M +2l)}{2n^2}
\end{equation*} 
This probability comes from the fact that we want to select any 0 in bin $\kappa$ in one of the parents and a 1 anywhere in the other parent. Obviously, as the number of 1s in both parents grows, this probability grows too. In Section~\ref{sec:exact} we also use the probability of failure:
\begin{equation*}
P_{\mathrm{F}}=1-P_{\mathrm{swap}}
\end{equation*} 

\subsection{Population and elitism assumptions}
We assume that each generation currently elite species in the population are distributed uniformly:  
\begin{equation*}
\alpha \sim \textrm{Uniform}\Big(\frac{1}{\mu}\Big)
\end{equation*}
This is a static model, i.e. this distribution does not change throughout the run of the algorithm; we will consider a dynamic model for this distribution in future work. We also assume that the rate of elitism (the number of species saved for the next generation) is $\textit{high enough}$, that is, high enough to keep all elite species. 

\section{Derivation of the expectation of convergence time}
We present three main results: exact, approximate and asymptotic. The latter two are necessary, since the complete one does not have a closed form.   

\subsection{Exact expression}
\label{sec:exact}

We start with introducing the probability of failure to improve $V_{\kappa}$:
\begin{equation}
P(G_{0})=\sum_{j=0}^{\frac{\lambda}{2}}P(G_{0}|H_j)\sum_{\alpha=1}^{\mu}P(H_j|\alpha)P(\alpha)
\label{eq:prob_fail1}
\end{equation}
where $H_j$ is $j$th elite pair in the recombination pool $\lambda$, and $\alpha$ is the number of elite species in the population $\mu$. 

The probability to fail to improve a bit in a bin given $l$ improvements so far is:
\begin{align}
P(G_{0 l})&=\frac{1}{\mu}\sum_{j=0}^{\frac{\lambda}{2}}\Big(\frac{2n^2-(M-2l)(n+\kappa M+2l)}{2n^2}\Big)^j \binom{\frac{\lambda}{2}}{j} \nonumber \\
&\cdot \sum_{\alpha=1}^{\mu}\Big(\frac{(\alpha(\alpha+2 \mu(\mu-\alpha)))^2}{\mu^4}\Big)^j \Big(1-\frac{(\alpha(\alpha+2 \mu(\mu-\alpha)))^2}{\mu^4}\Big)^{\frac{\lambda}{2}-j} \nonumber \\
&=\frac{1}{\mu}\sum_{j=0}^{\frac{\lambda}{2}}P_{\mathrm{F}}^{j}\binom{\frac{\lambda}{2}}{j}\sum_{\alpha=1}^{\mu}(P_{\mathrm{sel}}(\alpha))^j(1-P_{\mathrm{sel}}(\alpha))^{\frac{\lambda}{2}-j} \nonumber \\
&=\frac{1}{\mu}\sum_{\alpha=1}^{\mu} \sum_{j=0}^{\frac{\lambda}{2}} \binom{\frac{\lambda}{2}}{j}(P_{\mathrm{F}}P_{\mathrm{sel}}(\alpha))^{j} (1-P_{\mathrm{sel}}(\alpha))^{\frac{\lambda}{2}-j} \nonumber \\
&=\frac{1}{\mu}\sum_{\alpha=1}^{\mu}(1-P_{\mathrm{sel}}(\alpha)P_{\mathrm{swap}})^{\frac{\lambda}{2}}
\end{align}

The last step is due to the Binomial expansion: $\sum_{k=0}^{n}\binom{n}{k}a^{k}b^{n-k}=(a+b)^n$. Therefore, the probability of an increase in the auxiliary function is:
\begin{equation*}
P(G_l)=1-P(G_{0l})=1-\frac{1}{\mu}\sum_{\alpha=1}^{\mu}(1-P_{\mathrm{sel}}(\alpha)P_{\mathrm{swap}})^{\frac{\lambda}{2}}
\end{equation*}

The expected time until the next improvement of the auxiliary function of a bin $\kappa$ is: 
\begin{equation}
\mathbf{E}T_{\kappa}=\sum_{l=0}^{\frac{M}{2}-1}\frac{1}{P(G_l)}
\end{equation}
and, finally, summing over all $\kappa$ from 1 to $K$ we obtain (since $G$ depends on both $l$ and $\kappa$):
\begin{align}
\mathbf{E}\tau_{(\mu+\lambda)EA_{1BS}}&=\sum_{\kappa=1}^{K}\sum_{l=0}^{\frac{M}{2}-1}\frac{1}{P(G_{l,\kappa})} \nonumber \\
&=\sum_{\kappa=1}^{K}\sum_{l=0}^{\frac{M}{2}-1}\frac{1}{1-\frac{1}{\mu}\sum_{\alpha=1}^{\mu}(1-(\alpha(2\mu-\alpha))^2 P_{swap})^{\frac{\lambda}{2}}} \nonumber \\
&=\sum_{\kappa=1}^{K}\sum_{l=0}^{\frac{M}{2}-1}\frac{1}{1-\frac{1}{\mu}\sum_{\alpha=1}^{\mu}(1-\frac{(\alpha((2 \mu-\alpha)))^2(M-2l)(n+\kappa M+2l)}{2\mu^4 n^2})^{\frac{\lambda}{2}}}
\label{eq:rr_1bs_unif}
\end{align}
The benefit of  larger population sizes is clear in the $\frac{1}{\mu}$ term in front of the sums in the denominator. Also, increases in the size of $\lambda$ lead to reductions in the probability of failure.

We test this expression numerically for different values of $n, \mu, \lambda$ (see Appendix C). Unfortunately, this expression does not seem to exist in closed form, so we instead go ahead with finding an approximation to it in the next subsection. 

\subsection{Approximate and asymptotic expressions}
\begin{align}
P(G_{0l})&=\frac{1}{\mu}\sum_{\alpha=1}^{\mu}\Big(1-\frac{(\alpha(\alpha+2(\mu-\alpha)))^2 (M-2l)(n+\kappa M+2l)}{2 \mu^4 n^2}\Big)^{\frac{\lambda}{2}} \nonumber \\
&= \frac{1}{\mu}\sum_{\alpha=1}^{\mu}\Big(1-\frac{(\alpha(2 \mu-\alpha))^2(M-2l)(n+\kappa M+2l)}{2 \mu^4 n^2}\Big)^{\frac{\lambda}{2}} \nonumber \\
&\approx \frac{1}{\mu} \sum_{\alpha=1}^{\mu} e^{-\frac{\lambda(\alpha(2 \mu-\alpha))^2(M-2l)(n+\kappa M+2l)}{4 \mu^4 n^2}} \approx \frac{1}{\mu} \int_{1}^{\mu}e^{-\Big(\frac{\alpha(2 \mu-\alpha)}{\sqrt{\gamma}}\Big)^2} d \alpha
\label{eq:int1}
\end{align}
The last step in the summand was due to $\lim_{n \to \infty}(1-\frac{1}{n})=\frac{1}{e}$. Note that $\gamma=\frac{4 \mu^4 n^2}{\lambda(M-2l)(n+\kappa M+2l)}$, and, assuming that $\mu=\lambda$, the upper bound on $\gamma$ is $\frac{4 \mu^4 n^2}{\lambda(M-2l)(n+kM+2l)} < \frac{4 \mu^4 n^2}{3 \lambda n}=O(\mu^{3}n)$. although for monotonically decreasing functions, such as this one, by the integral test the sum is larger than the corresponding integral; for $\mu << n$ the sum is closely approximated by the integral. 

We denote $I_{1}=\int_{1}^{\mu} f(\alpha)d \alpha = \int_{1}^{\mu}e^{-\Big(\frac{\alpha(2 \mu-\alpha)}{\sqrt{\gamma}}\Big)^2} d \alpha$. Expanding the function inside the integral as a Taylor series around $\alpha_{0}=1$ up to the second term, we get (since $f'(\alpha_{0})=-\frac{4(2 \mu-1)(\mu-1)}{e^{(\frac{2 \mu-1}{\sqrt{\gamma}})^{2}} \gamma}$):
\begin{equation}
f(\alpha) \approx e^{-(\frac{2 \mu-1}{\sqrt{\gamma}})^2}-\frac{4(2 \mu-1)(\mu-1)(\alpha-1)}{e^{(\frac{2 \mu-1}{\sqrt{\gamma}})^{2}} \gamma}  
\end{equation}

Therefore, the integral turns into:
\begin{align}
I_{1}&=\int_{1}^{\mu} f(\alpha)d \alpha \approx \int_{1}^{\mu} \Bigg( e^{-(\frac{2 \mu-1}{\sqrt{\gamma}})^2}-\frac{4(2 \mu-1)(\mu-1)(\alpha-1)}{e^{(\frac{2 \mu-1}{\sqrt{\gamma}})^{2}} \gamma} \Bigg) d \alpha \nonumber \\
&= e^{-(\frac{2 \mu-1}{\sqrt{\gamma}})^2}(\mu-1)\Big[1-\frac{2(2 \mu-1)(\mu-1)^2}{\gamma} \Big]
\label{eq:int11}
\end{align}

The probability of failure is approximately (with the assumptions specified above):
\begin{equation*}
P(G_{0l}) \approx \frac{e^{-(\frac{2 \mu-1}{\sqrt{\gamma}})^2}(\mu-1)\Big[1-\frac{2(2 \mu-1)(\mu-1)^2}{\gamma} \Big]}{\mu}
\end{equation*}

Accordingly, the probability of a successful swap is: 
\begin{equation*}
P(G_l) \approx 1-\frac{e^{-(\frac{2 \mu-1}{\sqrt{\gamma}})^2}(\mu-1)\Big[1-\frac{2(2 \mu-1)(\mu-1)^2}{\gamma} \Big]}{\mu}
\end{equation*}

Using the sum of expectations of Geometric random variables with different parameters, the expected time until filling a bin, i.e. improvement of the fitness function, is: 
\begin{align}
\mathbf{E}T_{\kappa}=\sum_{l=0}^{\frac{M}{2}-1}\frac{\gamma}{\gamma-\gamma e^{-(\frac{2 \mu-1}{\sqrt{\gamma}})^2} + 2(2 \mu-1)(\mu-1)^2 e^{-(\frac{2 \mu-1}{\sqrt{\gamma}})^2}}
\end{align}  

We make two approximations here. First, we use a Riemanian sums approximation to obtain $[0,1]$ bounds on the integral, and then expand the integrand in Taylor series with 2 terms around the midpoint to obtain a good approximation of the integral. The Riemanian sums approximation is defined by:
\begin{equation*}
\lim_{n \to \infty} \sum_{j=1}^{n}f(x_j) =n \int_{0}^{1}f(nx)dx +o(n) 
\end{equation*}  
and $\gamma$ is transformed accordingly:
\begin{equation*}
\gamma=\frac{4 \mu^4 n^2}{(M-2(\frac{M}{2}-1)l)(n+\kappa M+(\frac{M}{2}-1)l)}
\end{equation*}

Then:
\begin{equation}
I_2=\int_{0}^{1}\frac{\gamma dl}{\gamma-\gamma e^{-(\frac{2 \mu-1}{\sqrt{\gamma}})^2} + 2(2 \mu-1)(\mu-1)^2 e^{-(\frac{2 \mu-1}{\sqrt{\gamma}})^2}}
\label{eq:integral2}
\end{equation}

Therefore, the expected first hitting time until the evolution of the bin $\kappa$ is (the rather long Taylor series expansion of the integrand is given in the Appendix A):
\begin{align}
\mathbf{E}T_{\kappa} &\approx \Big(\frac{M}{2}-1\Big)\int_{0}^{1}\frac{\gamma dl}{\gamma-\gamma e^{-(\frac{2 \mu-1}{\sqrt{\gamma}})^2} + 2(2 \mu-1)(\mu-1)^2 e^{-(\frac{2 \mu-1}{\sqrt{\gamma}})^2}} \nonumber \\
&=\frac{4 \mu^4 n^2(\frac{M}{2}-1)}{\lambda(\frac{M}{2}+1)(\frac{M}{2}+n+\kappa M-1)\Big[\frac{2(2 \mu-1)(\mu-1)^2}{\sigma_1}+\frac{4 \mu^4 n^2}{\lambda(\frac{M}{2}+1)\sigma_2}-\frac{4 \mu^4 n^2}{\lambda(\frac{M}{2}+1)\sigma_2 \sigma_1} \Big]}
\label{eq:fht2}
\end{align}
where: 
\begin{equation*}
\sigma_1=e^{\frac{\lambda(2 \mu-1)^2 (\frac{M}{2}+1) \sigma_2}{4 \mu^4 n^2}}
\end{equation*}
\begin{equation*}
\sigma_2=\frac{M}{2}+n+\kappa M-1
\end{equation*}
Note that $\sigma_1$ has the interesting property (given $\mu=\lambda$) that:
\begin{equation*}
\lim_{n \to \infty} e^{\frac{\lambda(2 \mu-1)^2 (\frac{M}{2}+1)(\frac{M}{2}+n+\kappa M-1)}{4 \mu^4 n^2}}=\lim_{n \to \infty}e^{\frac{M(M+n+KM)}{\mu n^2}}=\lim_{n \to \infty} e^{\frac{M}{\mu n}+O\Big(\frac{M^2}{\mu n^2}\Big)}=1
\end{equation*}
which means, that for sufficiently large values of $n$ and $\mu$ the second and the third terms in the square brackets cancel each other out, and the first term is just $2(2 \mu-1)(\mu-1)^2$.

Finally, summing over all $\kappa$, the number of bins in the string, we get the approximation of the convergence time of the $(\mu+\lambda)$ algorithm on RR test function:
\begin{align}
\mathbf{E}\tau^{RR}_{(\mu+\lambda)EA_{1BS}}& \approx \frac{2 \mu^4 n^2(M-2)}{\lambda(M+2)(2 \mu-1)(\mu-1)^2}\sum_{\kappa=0}^{K-1}\frac{1}{\frac{M}{2}+n-1+\kappa M} \nonumber \\
&=\frac{2 \mu^4 n^2(M-2)}{\lambda M(M+2)(2 \mu-1)(\mu-1)^2}\Big[\psi_{0}\Big(\frac{\frac{M}{2}+n-1+M+KM}{M}\Big) - \psi_{0}\Big(\frac{\frac{M}{2}+n-1}{M}\Big) \Big] \nonumber \\
&\approx \frac{2 \mu^4 n^2(M-2)}{\lambda M(M+2)(2 \mu-1)(\mu-1)^2}\log \Big(1+\frac{2KM}{M+2n}\Big)
\label{eq:expect2}
\end{align}
where $\psi_{0}$ is a Digamma function (see e.g. \cite{abramowitz, knuth95}). In the derivation of the asymptotic expression for this bound,  all population-related terms cancel out (since $\mu=\lambda$ and both numerator and denominator have the highest term $\mu^4$), and the order of convergence is 
\begin{equation}
\mathbf{E}\tau^{RR}_{(\mu+\lambda)EA_{1BS}}=O\Bigg(\frac{n^2 \log\Big(1+\frac{KM}{M+n} \Big)}{M}\Bigg)
\label{eq:expect3}
\end{equation}
which seems to be a comparable result compared to those available in literature covering fitness functions with plateaus of fitness (e.g. \cite{heyao02, chenhe09, storchwegener03}).
\section{Conclusions and Future Work}
We have derived three expressions for convergence of an elitist $(\mu+\lambda)$EA$_{1BS}$ on Royal Roads test function: exact, approximate and asymptotic. Although the exact expression for the expected convergence time clearly shows the benefit of increase in the population (at least when the population is relatively small), the approximate result has an equal order of population  and asymptotic has none at all due to cancellation.

An important assumption for the approximation of $\mathbf{E}\tau^{RR}_{(\mu+\lambda)EA_{1BS}}$ was that $\mu << n$, but we never specified the relation, unlike in \cite{chen11}. This is something to look at in the future. Since the effect of the population is known to be problem-specific, we will be able to get good insights into it for unimodal functions with plateaus, such as Royal Roads.

We have performed our analysis assuming Uniform distribution of elite species in the population, something noone seems to have done in EA literature before. This is a static approach to convergence (i.e. the distribution assumption does not change throughout the run of the algorithm). We would like to look at the dynamics of the elite species and their effect on the probability of success, $P(G_l)$ and expected convergence time.     
\bibliographystyle{alpha}
\bibliography{mybib7}
\appendix
\section{Taylor series approximation of the integrand}
We give the expression for the Equation \ref{eq:fht2} here due to its length. It's Taylor series expansion of the integrand around midpoint of the interval (0.5)
\begin{align*}
\phi(l) &\approx \frac{4 \mu^2 n^2}{\lambda(\frac{M}{2}+1)\sigma_{1} \sigma_{3}} + s_{3} \Bigg( \frac{s_{3}\Big\{\frac{\frac{s_1}{\sigma_{2} \sigma_{4}}-\frac{s_2}{\mu^4 n^2 \sigma_{2}(\frac{M}{2}+1)}}{\sigma_{3}}-\frac{16(M-2)}{\sigma_2 \sigma_{3}^2(\frac{M}{2}+1)}\Big\}-s_{3}\frac{(16M-32) \sigma_5}{\sigma_{4}\sigma_{3}^2}+\varphi_{1}}{(\frac{M}{2}+1) \sigma_{1}^2 \sigma_3}+\frac{4(M-2) \sigma_5}{\sigma_{1} \sigma_{3}^2 \sigma_{4}} \Bigg)\\ 
& \cdot (l-\frac{1}{2})
\end{align*} 
where 
\begin{equation*}
\sigma_{1}=\frac{2(\mu-1)(\mu-1)^2}{\sigma_2} + \frac{4 \mu^2 n^2}{\lambda(\frac{M}{2}+1)\sigma_3}-\frac{4 \mu^4 n^2}{\lambda \sigma_2 \sigma_3(\frac{M}{2}+1)}
\end{equation*}
\begin{equation*}
\sigma_2=e^{\frac{\lambda(2 \mu-1)(\frac{M}{2}+1) \sigma_3}{4 \mu^4 n^2}}, \ \sigma_{3}=\frac{M}{2}+n+kM-1, \ \sigma_{4}=\Big(\frac{M}{2}+1\Big)^2, \ \sigma_5=n+kM-2
\end{equation*}
\begin{equation*}
s_{1}=16(M-2), \ s_{2}=4 \lambda(2 \mu-1)^2((M-2)(\frac{M}{2}+1)-\sigma_{3}(M-2)), \ s_{3}=\frac{\mu^4 n^2}{\lambda}
\end{equation*}
\begin{equation*}
\varphi_{1}=\frac{2(2 \mu-1)^3 (\mu-1)^2 (2M+2n+2kM-nM-kM^2-4)}{s_{3}\sigma_{2}}
\end{equation*}
\section{Additional Derivation Details}
To derive expressions in Section \ref{sec:exact}, we extensively used properties of independent Geometric RVs that are not identically distributed, which is also known as Coupon collector's problem (see e.g. \cite{knuth95}): if $X_i \sim Geom(p_i)$ it expectation is $\mathbf{E}[X_i]=\frac{1}{p_i}$. Therefore, if $Y=\sum_{i=1}^{n}X_{i}, \ \mathbf{E}Y=\sum_{i=1}^{n} \mathbf{E}[X_{i}]=\frac{1}{p_1}+\frac{1}{p_2}+\ldots + \frac{1}{p_n}$.

For Equation \ref{eq:prob_fail1} we use the Law of total probability twice: first, conditioning on $H_j$, then on $\alpha$:
\begin{equation*}
P(A)=\sum_{i=1}^{m}P(A|B_i)P(B_i)=\sum_{i=1}^{m}P(A|B_i)\sum_{j=1}^{n}P(B_i|C_j)P(C_j)
\end{equation*}
\section{Numerical results to verify Equation \ref{eq:rr_1bs_unif}}
Column $\tilde{\tau}_{(\mu+\lambda)EA_{1BS}}$ was obtained by running the algorithm with different parameters 20 times, each run was 2000 generation each. The earliest achievement of the global minimum for each run was saved and then averaged over. 
\begin{table}
\centering
\begin{tabular}{|c|c|c|c|c|c|c|}
\hline
$n$&K&M&$\mu$&$\lambda$&$\mathbf{E}\tau^{RR}_{(\mu+\lambda)EA_{1BS}}$&$\tilde{\tau}_{(\mu+\lambda)EA_{1BS}}$\\
\hline
\multirow{4}{*}{32}&\multirow{4}{*}{4}&\multirow{4}{*}{8}&4&4&145&315.31\\
&&&10&10&72.4&268.22\\
&&&20&20&44.2&192.29\\
&&&30&30&34.5&173.56\\
\hline
\multirow{4}{*}{64}&\multirow{4}{*}{8}&\multirow{4}{*}{8}&4&4&570.62&612.46\\
&&&10&10&279.88&497.93\\
&&&20&20&153.46&454.47\\
&&&30&30&112.30&372.04\\
\hline
\multirow{4}{*}{128}&\multirow{4}{*}{16}&\multirow{4}{*}{8}&4&4&2264.36&1365\\
&&&10&10&1048&1239\\
&&&20&20&570.44&1091.5\\
&&&30&30&401.99&949.4\\
\hline
\end{tabular}
\caption{Theoretical and computational bounds for $(\mu+\lambda)EA^{RR}_{1BS}$} 
\end{table}
\end{document}